\documentclass[twoside,11pt]{article}
\usepackage{jair, theapa, rawfonts}
\usepackage{booktabs,array}

\usepackage{amsfonts}

\usepackage{algorithmic}
\usepackage{graphicx}
\usepackage{textcomp}
\usepackage{amsmath}
\DeclareMathOperator*{\argmax}{argmax}
\usepackage{subcaption}

\jairheading{1}{2021}{1-15}{7/21}{-/21}
\ShortHeadings{An Analysis of Reinforcement Learning for Malaria Control}
{Makondo, Folarin, Zitha, \& Remy}
\firstpageno{1}

\begin{document}

\title{An Analysis of Reinforcement Learning for Malaria Control}

\author{\name Ndivhuwo Makondo \email ndivhuwo.makondo@ibm.com \\
       \addr IBM Research-Africa, Johannesburg,\\
       South Africa\\
       \addr School of Computer Science and Applied Mathematics,\\
       University of the Witwatersrand, Johannesburg,\\ 
       South Africa
       \AND
       \name Arinze L. Folarin \email arinze@aims.ac.za \\
       \addr African Institute of Mathematical Sciences (AIMS),\\ 
       Cape Town, South Africa
       \AND
       \name Simphiwe Zitha \email nhlanhla34@gmail.com \\
       \addr IBM Research-Africa, Johannesburg,\\
       South Africa\\
       \addr School of Computer Science and Applied Mathematics,\\
       University of the Witwatersrand, Johannesburg,\\ 
       South Africa
       \AND
       \name Sekou L. Remy \email sekou@ke.ibm.com \\
       \addr IBM Research-Africa, Nairobi,\\
       Kenya}


\maketitle

\begin{abstract}
Previous work on policy learning for Malaria control has often formulated the problem as an optimization problem assuming the objective function and the search space have a specific structure. 
The problem has been formulated as multi-armed bandits, contextual bandits and a Markov Decision Process in isolation. Furthermore, an emphasis is put on developing new algorithms specific to an instance of Malaria control, while ignoring a plethora of simpler and general algorithms in the literature. 
In this work, we formally study the formulation of Malaria control and present a comprehensive analysis of several formulations used in the literature. 
In addition, we implement and analyze several reinforcement learning algorithms in all formulations and compare them to black box optimization. 
In contrast to previous work, our results show that simple algorithms based on Upper Confidence Bounds are sufficient for learning good Malaria policies, and tend to outperform their more advanced counterparts on the malaria OpenAI Gym environment.
\end{abstract}

\section{Introduction}
\label{sec:introduction}

Malaria is a life-threatening disease caused by parasites that are transmitted to people through the bites of infected female Anopheles mosquitoes \cite{world2018world}. 
It is one of the leading causes of morbidity and mortality in Africa, accounting for approximately 90\% illness cases and 91\% death cases in 2018 worldwide \cite{world2018world}. 
With the explosion of research in AI, there has been a growing research interest in developing Malaria control precaution (Malaria intervention) tools to improve the disease diagnosis and management \cite{korenromp2016malaria}. 
The aim of scaling up Malaria prevention and control interventions is to reduce the mortality and morbidity rate of the disease \cite{bent2018novel}. 

Currently, human decision makers (governments, NGOs and charities) spend an extensive amount of time making policies that decide which Malaria intervention actions are to be explored for different sub-populations \cite{bent2018novel}.
However, due to the complexity of the space of possible actions for Malaria intervention, it becomes inefficient for human decision makers to explore different actions with high return \cite{bent2018novel}.
Therefore, it is crucial to develop artificial intelligence (AI) driven decision support systems that will address such inefficiencies, thereby maximizing the effectiveness of malaria intervention.
In this work, we are interested in learning about the formulation of Malaria control interventions and understanding the performance of different classes of AI algorithms, particularly looking at the interventions using insecticide-treated nets (ITNs) and indoor residual spraying (IRS), which have been shown to be effective in sub-Saharan Africa (SSA) \cite{bhatt2015effect}. 
We study the problem as Multi-armed Bandits and contextual Bandits as well as a Markov Decision Process (MDP) formulation.

Earlier work looking at computational tools for Malaria control investigated the use of mathematical models of malaria epidemiology for providing realistic predictions of likely epidemiological outcomes based on existing control strategies. 
In \cite{stuckey2012simulation}, stochastic simulation models of malaria using the OpenMalaria \cite{smith_et_al_2008} platform were calibrated to data from ongoing field studies and literature for the Rachuonyo district in Western Kenya.
Following this work, \cite{bent2018novel} formulated malaria control as a Multi-armed Bandit problem and implemented three AI algorithms based on Genetic Algorithms, Batched Policy Gradients and Gaussian Process Upper/Lower Confidence Bounds (GP-ULCB). 
Their results showed that the three algorithms were able to find cost effective policies, where the policy space is made up of ITNs and IRS each year for 5 years.  

Recently, an OpenAI Gym environment for Malaria control was published \cite{pmlr-v123-remy20a}, which led to the emergence of work applying reinforcement learning (RL) to malaria control. 
In \cite{nguyen2019policy}, Q-learning was modified to exploit the break in the sequential structure of the gym model\footnote{The break may be an artefact of the instance of the Malaria model used in the gym environment, which may not apply to all malaria models} and was shown to outperform Bayesian optimization and genetic algorithm. 
In \cite{ali_optimal_2020}, Q-learning, SARSA and Deep Deterministic Deep Policy Gradients (DDPG) were compared and DDPG was shown to perform better. 
While these previous works provide evidence that RL can learn good malaria policies, they ignore a plethora of much simpler and more general RL algorithms.
Furthermore, while some (e.g., \cite{nguyen2019policy}) focus on developing new algorithms for a specific instance of the malaria environment (i.e., the gym environment \cite{pmlr-v123-remy20a}), they do not provide a comprehensive analysis of different algorithms that could apply to general models of malaria.

\textbf{Contributions} We study the formulation of Malaria control and provide a comprehensive evaluation of various formulations and different classes of AI algorithms on these formulations.
This provides insight into how different formulations of the problem limit the application of certain classes of algorithms and affect the performance of such algorithms.
Previous work in the literature do not provide a comprehensive analysis across different classes of algorithms and formulations. 
We show that simple RL algorithms in contextual and MDP formulations are capable of learning good Malaria policies using the gym environment \cite{pmlr-v123-remy20a} as a testbed.

This paper is organized as follows. Section \ref{related_work} briefly reviews application of machine learning and reinforcement learning in healthcare and epidemiology. 
Section \ref{malaria_form} formally introduces the Malaria control problem and describes the formulations used in the literature and analyzed in this paper. 
Section \ref{algorithms} presents the algorithms used in this paper and their applicability to the formulations in Section \ref{malaria_form}.
We provide our experimental analysis of the algorithms in each formulation and across formulations in Section \ref{experiments} and conclude the paper in Section \ref{conclusion}.

\section{Related Work}
\label{related_work}

Fuelled by the (rapid) increases in costs, complexity, the myriad of treatment options and the increasing availability of multimodal data, the application of AI and Machine Learning has also significantly increased in the recent past.
AI tools are increasingly used to uncover fundamental patterns that can be used to predict optimal treatments, minimize side effects, and reduce cost(s) and medical errors \cite{bennett2013artificial}. 
According to a review \cite{yu2019reinforcement}, AI has been applied in several healthcare domains, including dynamic treatment recommendation/regimes (DTR), automated medical diagnosis, health resource allocation and scheduling, optimal process control, drug discovery and development.

While both supervised learning (SL) and reinforcement learning (RL) techniques are applied in various healthcare domains, reinforcement learning has seen a significant increase in use recently \cite{wang2018supervised}. 
This is due to its ability to model sequential problems with delayed outcomes and uncertainty.
The majority of research using RL in healthcare is in dynamic treatment regimes, where the goal is to develop effective treatment regimes that can dynamically adapt to the varying clinical states and improve the long-term outcomes for patients \cite{yu2019reinforcement}. 
This includes DTR for diseases such as cancer \cite{zhao2009reinforcement,liu2017deep}, diabetes \cite{daskalaki2010preliminary,bothe2013use,daskalaki2013actor}, anemia \cite{malof2011optimizing,escandell2014optimization}, HIV \cite{parbhoo2014reinforcement,parbhoo2017combining,yu2019incorporating}, mental illnesses \cite{paredes2014poptherapy,pineau2009treating}, and DTR in critical care  \cite{weng2017representation,petersen2018precision}.

In addition, RL is applied in both observational settings, where historical data is used to train off-policy RL \cite{tseng2017deep,gottesman2018evaluating}, and simulation settings, where an RL algorithm interacts with a computational model of disease progression \cite{zhao2009reinforcement,ngo2018reinforcement}. 
Our work on Malaria control is related to DTR with simulated models, where we use an epidemiological model of malaria to find an optimal sequence of interventions.  

Due to the recent outbreak of the Covid-19 pandemic, there has been a plethora of work in using RL for finding optimal interventions such as lockdown levels and vaccination strategies.
Simulation models based on susceptible-infectious-recovered (SIR) model and its extension, the SEIR, were developed, and in contrast to sub-optimal, manually designed strategies from the epidemiological community, RL algorithms are used to interact with these models to find optimal interventions \cite{kwak2021deep,colas2020epidemioptim,ohi2020exploring,padmanabhan2021reinforcement}. 
This also extends to the epidemic control of other diseases such as mouth-and-foot \cite{probert2019context}, pandemic influenza \cite{libin2020deep}, and in our case, malaria \cite{bent2018novel,ali_optimal_2020,nguyen2019policy,wachira2020platform}.

\section{Malaria Control Formulation}
\label{malaria_form}

The OpenMalaria platform \cite{smith_et_al_2008} provides a simulation environment for an agent to learn \textit{optimal} policies for the control of malaria from epidemiological models grounded in real-world data. 
Each policy is associated with a cost that is modelled through an economic analysis of the stochastic simulation output, using calculations as specified in the health economics literature.
This provides a pathway to generate findings on benefit-cost analyses of considered policies as assessed by a variety of methods.
This includes Disability adjusted life years (DALYs), Simulated Costs and Cost effectiveness (See \cite{bent2018novel,briet2013effects} for a detailed description). 
A policy which meets pre-specified targets at the lowest cost could be considered to be the optimal policy. 

We generally pose searching for an optimal Malaria policy as an optimization problem 
\begin{equation}
    {\max}_{\mathbf{x} \in M} f(\mathbf{x}),
    \label{malaria_problem}
\end{equation}
where  $\mathbf{x} \in M \subset \mathbb{R}^d$ is a $d$-dimensional input space. 
We define $\mathbf{x}_{i}$ as the $i$th sample and $r_{i} = f(\mathbf{x}_{i}) + \epsilon_{i}$ as
a noisy observation of the objective function\footnote{Since we are maximizing, the objective function is the negative of the cost} at $\mathbf{x}_{i}$ and noise value $\epsilon_{i}$. 

\begin{figure}[!t]
\begin{center}
\centerline{\includegraphics[width=0.6\textwidth,clip]{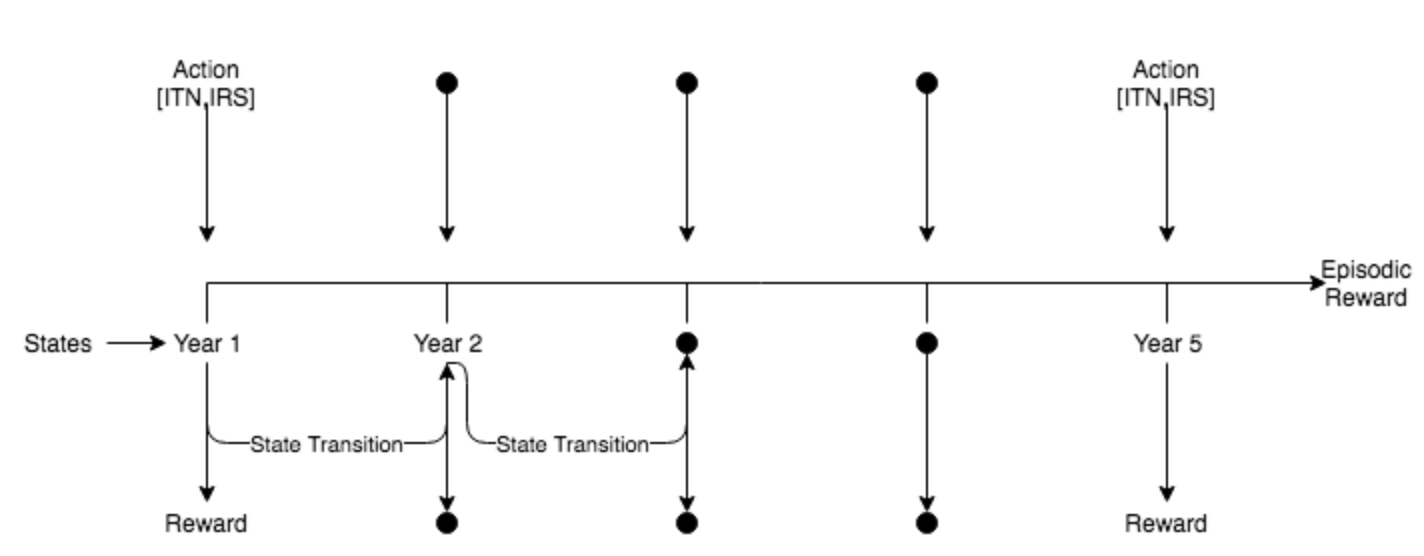}}
\caption{OpenAI Gym Malaria simulation \cite{pmlr-v123-remy20a}. The 2-dimensional action $[ITN,IRS]$ is applied each year and triggers a transition into the next year. The environment also responds with a numeric reward after each action is executed.}
\label{MalariaGym}
\end{center}
\end{figure}

An OpenAI Gym simulation environment for Malaria control \cite{pmlr-v123-remy20a} was recently developed as a benchmark for analyzing the performance of various learning algorithms \cite{ali_optimal_2020,nguyen2019policy}. The environment, shown in Figure \ref{MalariaGym}, consists of a simulated population of individuals over a 5-year intervention time frame. The objective is to find an optimal sequence of interventions over the years in steps of 1 year (5 time steps per episode). Each intervention is made up of a 2-dimensional, continuous action $\mathbf{a}_t \in A = \{a_{ITN},a_{IRS}\}$, where $a_{ITN},a_{ITN} \in [0,1]$. 

The action $a_{ITN}$ corresponds to the proportion of population coverage for the mass-distribution of long-lasting insecticide-treated nets (ITNs), and $a_{IRS}$ corresponds to the coverage of the population in which Indoor Residual Spraying (IRS) with pyrethroids is deployed. This environment models \eqref{malaria_problem} as a sequential decision making problem.

As such, we formally define a Malaria policy $\pi_i(s): S \rightarrow A$ as a mapping that produces a sequence $[(s_1,\mathbf{a}_1),...,(s_5,\mathbf{a}_5)]$, where $s_t$ ($t \in \{1,...,5\}$) is the state of the environment (corresponding to the year). We seek the optimal policy $\pi_{\star}$
\begin{equation}
    \pi_{\star} = {\argmax}_{\pi \in \Pi} R(\pi),
    \label{optimal_policy}
\end{equation}
from a sequence of noisy observations $D_{1:i} = \{\pi_{1:i},r_{1:i}\}$, where $r_i = R(\pi_i)$ is the observed stochastic reward returned by the environment and $\Pi$ is the space of all possible policies. Using the general optimization problem \eqref{malaria_problem}, the input vector $\mathbf{x}_i = [a_{i,1},a_{i,2},...,a_{5,1},a_{5,2}]$ is equivalent to the policy $\pi_i$, without the sequential structure.

The Malaria environment \cite{pmlr-v123-remy20a} returns immediate rewards $R_t(\mathbf{a}_t)$ after executing action $\mathbf{a}_t$ at time $t$, which sum to the episodic reward $r_i = R(\pi_i) = \sum_{t=1}^{5} R_t(\mathbf{a}_t) + \epsilon_i$. In \eqref{malaria_problem} the true, unobserved objective function $f(\mathbf{x}_i) = \sum_{t=1}^{5} R_t(\mathbf{a}_t)$ where the input vector $\mathbf{x}_i$ is converted to its equivalent policy $\pi_i$ to obtain the actions $\mathbf{a}_t$.

\textbf{Discrete actions} Some algorithms analyzed here assume a discrete action space. One solution is to discretize each dimension, $d$, of the action space. This is not problematic to our low-dimensional action space and has been used in healthcare domains \cite{raghu2017continuous,komorowski2018artificial}. However, one could alternatively apply algorithms for continuous action spaces \cite{gaskett1999q,krishnamurthy2019contextual}. With discretization we obtain a set of discrete actions 

\begin{equation}
    A_d(k) = (\frac{j}{k-1})^{k-1}_{j=0},
    \label{discrete_actions}
\end{equation}

where $k$ is the number of discrete actions in each dimension. $k = 11$ results in 11 discrete actions in each dimension in the range $[0.0,0.1,0.2,...,1.0]$, so the new representation of the action space is a 1-dimensional space, which contains 121 discrete actions in total (i.e., $\mathbf{a}_t = (a_{ITN},a_{IRS})_j$ where $j \in \{0,1,..,120\}$). The new discrete representation reduces the search space for smaller $k$'s but at the cost of losing accuracy on the optimal actions.

In this work we analyze several learning algorithms for Malaria control and categorize them based on the assumptions they make on the reward distribution $R(\pi)$ (or $f(\mathbf{x})$) and the nature of the problem. This includes reinforcement learning (RL) algorithms that assume \eqref{malaria_problem} is a sequential decision making problem (and thus use \eqref{optimal_policy}); and black box optimization algorithms that assume no structure on $f$. We analyze these algorithms on three formulations: 

\begin{itemize}
    \item Multi-armed bandit (Section \ref{context-free-formulation}), referred to as context-free bandit in this paper,
    \item Contextual bandit (Section \ref{contextual-formulation}),
    \item Markov Decision Process (MDP) (Section \ref{mdp_form}).
\end{itemize}

\subsection{Context-free Bandit}
\label{context-free-formulation}

\begin{figure}[!t]
\begin{center}
\centerline{\includegraphics[width=0.6\textwidth,clip]{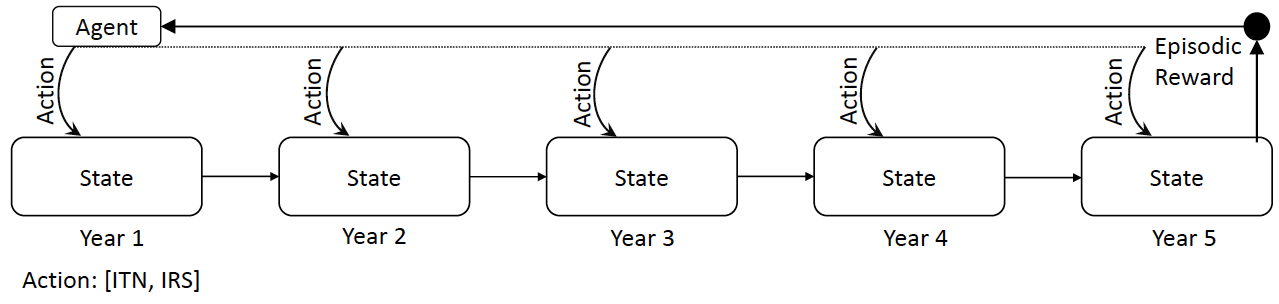}}
\caption{Context-free Malaria formulation.}
\label{context-free-form}
\end{center}
\end{figure}

The context-free formulation assumes there is a single state and models the policy $\pi_i(s)$ as a mapping to a single action $\mathbf{a}_i$ that is repeatedly applied every year. It makes no assumption about the reward distribution in each year and is only concerned with the final episodic reward $r_i$. Algorithms in this setting sequentially collect $D_{1:i} = \{\pi_{1:i},r_{1:i}\}$ to learn from until termination after $N$ episodes. Example applications of this formulation include developing models which may aid in the design of clinical trials to find the best treatment and ensuring more patients benefit from the treatment \cite{kuleshov2014algorithms}, and was also used for Malaria control \cite{bent2018novel}.

Figure \ref{context-free-form} shows the context-free setting for Malaria control. There is a single agent that chooses an action at the start of episode $i$, which is repeatedly applied each  year and the episodic reward $r_i$ is returned to the agent at the end of the episode. 

\subsection{Contextual Bandit}
\label{contextual-formulation}

While the context-free formulation involves a single state, the contextual bandit setting extends this to multiple states (or contexts) and involves both trial-and-error learning to search for the best policy, and association of each action with the states in which they are best. However, the policy in this setting ignores correlations between states, assuming the action in each state is independent of other states. An example use of this formulation is news article recommendation, where the goal is to recommend some news articles for a particular user (context) \cite{Li_etal_2010}; matching stress management interventions to individuals and their temporal circumstances over time \cite{paredes2014poptherapy}; and Malaria control \cite{nguyen2019policy}.

\begin{figure}[!t]
\begin{center}
\centerline{\includegraphics[width=0.6\textwidth,clip]{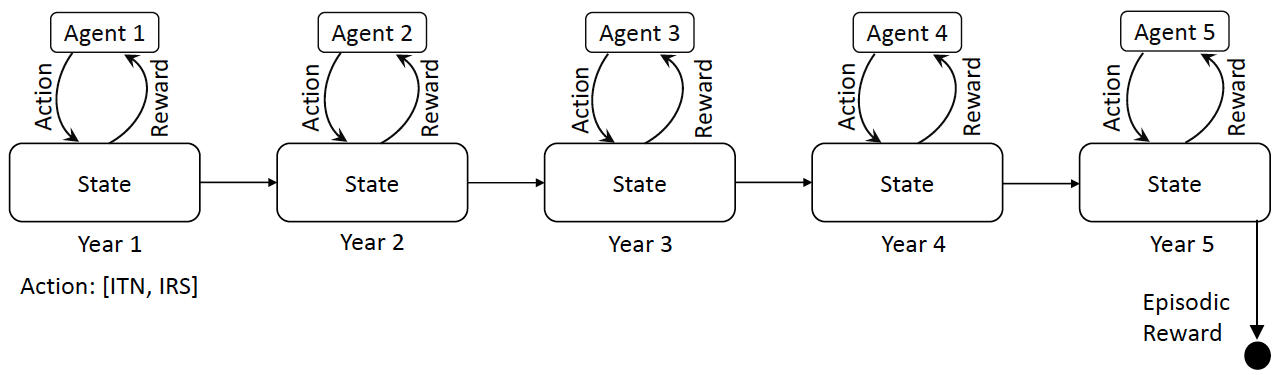}}
\caption{Contextual Malaria formulation \cite{ali_optimal_2020}}
\label{contextual-form}
\end{center}
\end{figure}

Figure \ref{contextual-form} shows the contextual bandit formulation for Malaria control, where there is an agent for each year that learns from immediate rewards $R_t(\mathbf{a}_t)$. The policy $\pi_i(s)$ thus sequentially learns actions for each state independently from observations $D^t_{1:i} = \{\mathbf{a}_{1:i},R_{1:i}\}^t$. The contexts are always provided in a sequence from year 1 to 5, as the environment always transitions in that manner independent of actions.

\subsection{Markov Decision Process}
\label{mdp_form}

The MDP formulation extends the contextual bandits to the formulation where the environment responds to actions and presents new situations to the agent. More concretely, in an episodic setting\footnote{We assume an episodic setting since the Malaria simulation always terminates at year 5.}, the agent and environment interact at each of a sequence of discrete time steps $t$. At each time step $t$, the agent receives the state of the environment, $s_t \in \mathbb{S}$, and on that basis selects an action $\mathbf{a}_t \in A$. One time step later, in part as a consequence of its action, the agent receives a numerical reward, $r_{t+1} \in \Re \subset \mathbb{R}$, and finds itself in a new state, $s_{t+1}$.

Formally, an MDP can be formulated as a tuple $(\mathbb{S},A,P_{a},R_{a},\lambda)$, where $\mathbb{S}$ is a set of states, $\mathbb{A}$ is a set of actions, $P_a(s,s^{\prime}) = Pr(s_{t+1} = s^{\prime} | s_{t} = s, a_{t} = a)$ is the transition function that defines the probability that action $\mathbf{a}$ in state $s$ will lead to state $s^{\prime}$ at time $t+1$, $R_{a}(s,s^{\prime})$ is the immediate reward received after transitioning from state $s$ to $s^{\prime}$, due to action $\mathbf{a}$; and $\lambda$ is the discount factor satisfying $0 \leq \lambda \leq 1$, which is usually close to 1.

The policy $\pi(s)$ if found by maximizing the expected return 
\begin{equation}
    G_t = \sum_{k=t+1}^{N} {\lambda}^{k-t-1}R_k,
    \label{mdp_return}
\end{equation}
which accumulates immediate rewards for each time step $t$ until the episode terminates. Since the state always transitions independent of actions, the transition function, $P_a$, is independent of the actions, i.e., $P_a = P_a(s,s^{\prime}) = Pr(s_{i+1} = s^{\prime} | s_i = s)$. Consequently, the reward distribution, $R_a$, depends on the current state and the previous action taken and no longer considers the previous state, i.e., $R_a(s^{\prime})$. The MDP formulations models correlations between states and is most widely used in healthcare applications \cite{bennett2013artificial,yu2019reinforcement}.

\section{Algorithms}
\label{algorithms}

This section describes algorithms analyzed in this paper and their application to Malaria control. They are categorized into RL and black box optimization, and the 3 settings applicable to them. Section \ref{rl_algorithms} describes central reinforcement learning algorithms, and Section \ref{optimization_baseline} describes black box optimization algorithms.   

\subsection{Reinforcement Learning}
\label{rl_algorithms}

This section describes central reinforcement learning algorithms categorized into value-based (Section \ref{rl_value}) and policy gradient (Section \ref{rl_policy}) methods in all applicable settings.

\subsubsection{Value-based Methods}
\label{rl_value}

Value-based methods help a learning agent to decide which action to select in order to achieve its goal of maximizing expected rewards by learning a value function of actions.

\textbf{Context-free setting} In the context-free setting, the true value, $Q_{\star}(\mathbf{a})$, of picking an action $\mathbf{a}$ is defined  as 

\begin{equation} 
 Q_{\star}(\mathbf{a})=\mathbb{E}[r_i|A_i=\mathbf{a}],
 \label{q_value}
\end{equation}

where $r_i$ is the reward of episode $i$ after selecting action $\mathbf{a}_i$. Once $Q_{\star}(\mathbf{a})$ is found, we can derive the optimal policy $\pi_{\star}(s)$ greedily: $\pi_{\star}(s) = {\argmax}_{\mathbf{a}}Q_{\star}(\mathbf{a})$. However, the learning agent has no access to $Q_{\star}$ and so needs to be estimated by executing an actions $\mathbf{a}_i = \pi_i(s)$ and observing rewards $r_i$, and the estimated action value, $Q(\mathbf{a})$, can be obtained incrementally from $D_{1:i} = \{\pi(s)_{1:i},r_{1:i}\}$:

\begin{equation} 
 Q_i(\mathbf{a})=Q_{i-1}(\mathbf{a}) + \alpha(r_i - Q_{i-1}(\mathbf{a})),
 \label{Q_update}
\end{equation}

where $\alpha \in (0,1]$ is the learning rate parameter.

An action selection strategy is required to collect $D_{1:i}$. It trades off exploration and exploitation, where exploration enables the discovery of new, potentially rewarding actions while exploitation uses the learned $Q_i(\mathbf{a})$ to choose the best known action. Several action selection strategies exist, with $\epsilon$-greedy and upper confidence bounds (UCB) being the most popular and widely used strategies. 

$\epsilon$-greedy is the combination of random exploration and the greedy strategy. The random strategy purely explores the environment whereas the greedy purely exploits the learned $Q_t(\mathbf{a})$ and is defined as 

\begin{equation} 
 A_i = {\argmax}_{\mathbf{a}}Q_i(\mathbf{a}),
 \label{greedy_eq}
\end{equation}
where the ${\arg\max}_{\mathbf{a}}$ function selects the action with maximum estimated value after episode $i$. The $\epsilon$-greedy works by selecting actions based on

\[
    A_i = 
\begin{cases}
    \argmax_{\mathbf{a}} Q_i(\mathbf{a}),&\text{with probability} \hspace{0.3cm} 1-\epsilon\\
    \text{a random action}, &\text{with probability} \hspace{0.3cm} \epsilon,
\end{cases}
\]

where $\epsilon$ is a hyperparameter that defines the probability of exploration.

The UCB algorithm is one of the efficient algorithms used in multi-armed bandit problems to address exploration-exploitation trade off. It selects actions by choosing the action that maximizes the sum of the exploitation and exploration terms:

\begin{equation}
    A_i = {\argmax}_{\mathbf{a}} Q_i(\mathbf{a}) + c\sqrt{\frac{\ln{i}}{N_i(\mathbf{a})}},
    \label{ucb_original}
\end{equation}

where $A_i$ is the action selected at episode $i$, $\argmax_{\mathbf{a}} Q_i(\mathbf{a})$ is the exploitation term and $c\sqrt{\frac{\ln{i}}{N_i(\mathbf{a})}}$ is the exploration term. $N_i(\mathbf{a})$ denotes the number of times that action $\mathbf{a}$ has been selected prior to episode $i$, and $c > 0$ is a hyperparameter that controls the degree of exploration. 

The action selection algorithms described above require a maximization of the action-value estimate, $Q_i(\mathbf{a})$, which is expensive for the continuous action space of the Malaria environment. We use the discretization scheme \eqref{discrete_actions}. When applied to the discretized action space, $Q_i(\mathbf{a})$ is a list of values for each action and each value of $\mathbf{a}$ is updated using \eqref{Q_update} when $\mathbf{a}$ is executed at episode $i$ and $r_i$ is observed.  

\textbf{Contextual setting} The value-based methods in the context-free setting can be easily generalized to the contextual setting by keeping separate action value functions for each context, which can be defined by the state-action value function $Q_i(s,\mathbf{a})$. In the descretized action space, $Q_i(s,\mathbf{a})$ is a matrix where the rows correspond to states and columns to the discrete actions. The update rule \eqref{Q_update} is thus generalized to

\begin{equation} 
 Q_i(s_t,\mathbf{a}_t)=Q_{i-1}(s_t,\mathbf{a}_t) + \alpha(R_t - Q_{i-1}(s_t,\mathbf{a}_t)),
 \label{Q_update_contexual}
\end{equation}
where the immediate reward $R_t$ is used in place of the episodic reward $r_i$.

\textbf{MDP setting} As discussed in Section \ref{mdp_form}, the MDP formulation extends the contextual formulation to situations where the environment responds to actions and presents new situations to the agent. Thus, the situations are correlated to each other. Q-learning is a popular valued-based method that models the action-value function \eqref{q_value} as a function of states and action

\begin{equation}
    Q_{\star}(s,\mathbf{a}) = \mathbb{E}[R_t|S_t=s,A_t=a],
    \label{q_function}
\end{equation}
which denotes the value of taking action $A_t$ from state $S_t$ following the optimal policy. The update rule \eqref{Q_update} is thus generalized to 

\begin{multline} 
 Q_i(s_t,\mathbf{a}_t)=Q_{i-1}(s_t,\mathbf{a}_t) + \alpha[R_t + \lambda \cdot \max_{\mathbf{a}} Q_{i-1}(s_{t+1},\mathbf{a}_t) - Q_{i-1}(s_t,\mathbf{a}_t)],
 \label{Q_func_update}
\end{multline}
where $\lambda$ is the discount factor and $\alpha$ the learning rate. The right term of \eqref{Q_func_update} is the temporal difference (TD) and incrementally approximates the return \eqref{mdp_return}, i.e., $G_t \approx R_t +  \lambda \cdot \max_{\mathbf{a}} Q_{i-1}(s_{t+1},\mathbf{a}_t) - Q_{i-1}(s_t,\mathbf{a}_t)$. 

The $\epsilon$-greedy algorithm is generally used for action selection. In this work, we also use UCB for action selection in the MDP settings. To this end, in addition to $Q(s,\mathbf{a})$ we also store $N(s,\mathbf{a})$, the number of times action $\mathbf{a}$ was selected in state $s$ and \eqref{ucb_original} is thus generalized to 

\begin{equation}
    A_t = {\argmax}_{\mathbf{a}} Q_i(s_t,\mathbf{a}_t) + c\sqrt{\frac{\ln{i}}{N_i(s_t,\mathbf{a}_t)}}.
    \label{td_cucb}
\end{equation}

On one hand, we can view this algorithm as Q-learning with UCB action selection, and on the other hand as a contextual UCB with correlated states and actions that is updated using TD learning. We therefore term this algorithm TD Contextual UCB (TD-CUCB). We refer to Q-learning with $\epsilon$-greedy as simply Q-learning.

\subsubsection{Policy Gradients} 
\label{rl_policy}

In contrast to value-based methods, policy gradients learn a numerical preference for each action $\mathbf{a}$ denoted as $H_i(\mathbf{a}_t)$ (contrast with \eqref{q_value}), which is converted into a probability distribution using a soft-max distribution. In the context-free setting, the policy is then a distribution

\begin{equation}
    \pi_i(\mathbf{a}) = \frac{e^{H_i(\mathbf{a})}}{\sum_{b=1}^{k} e^{H_i(\mathbf{a}_b)}} = Pr\{A_i = \mathbf{a}\},
    \label{pg_actions}
\end{equation}

where $\pi_i(\mathbf{a})$ is the probability of action $\mathbf{a}$ to be selected at episode $i$. The action preferences are initialized to 0 and updated every time step using stochastic gradient ascent

\begin{align}
    H_{i+1}(A_i) &= H_i(A_i) + \alpha(R_i - b)(1-\pi_i(\mathbf{a})),\nonumber\\
    \mbox{and }  H_{i+1}(\mathbf{a}_i) &= H_i(\mathbf{a}_i) + \alpha(R_i - b)\pi_i(\mathbf{a}), \mbox{ for all }  \mathbf{a} \neq A_t,
    \label{pg_update}
\end{align}

where $A_t$ is the selected action, $\alpha > 0$ is the learning rate and $b \in \mathbb{R}$ is a baseline. Here $b$ is computed as the average of all rewards received up to episode $i$. If the reward is higher than the baseline, then the probability of taking $A_t$ in the future is increased, and if the reward is below baseline, then probability is decreased. The non-selected actions move in the opposite direction.

Similar to value-based methods, we discretize the action space into 121 actions and learn a single policy $\pi(\mathbf{a})$ that predicts a one-shot action in the beginning of each episode. In the contextual setting policy $\pi_i(s,\mathbf{a})$ depends on the state and, similar to value-based methods, is updated independently for each state. 

We can generalize the action preferences to $h(s,\mathbf{a},\theta) \in \mathbb{R}$ for each state-action pair in the MDP setting, where $h$ is parameterized by $\theta$. \eqref{pg_actions}  is thus generalized to 
\begin{equation}
    \pi_i(\mathbf{a}|s,\theta) = \frac{e^{h(s,\mathbf{a},\theta)}}{\sum_{b=1}^{k} e^{h(s,\mathbf{a}_b,\theta)}}.
    \label{pg_actions_mdp}
\end{equation}

We parameterize the action preferences $h(s,\mathbf{a},\theta)$ using a neural network, where $\theta$ is the vector of all the connection weights of the network. In this work we consider the simplest policy gradient method, REINFORCE, which has the following update rule:

\begin{equation}
    \theta_{t+1} = \theta_t + \alpha(G_t - b(s_t))\frac{\Delta_{\theta}\pi(\mathbf{a}_t|s_t,\theta_t)}{\pi(\mathbf{a}_t|s_t,\theta_t)},
    \label{reinforce_update}
\end{equation}
where $G_t$ is the expected return \eqref{mdp_return} and $b(s)$ is now a state-dependent baseline. We normalize rewards received up to time $t$ and use this as the baseline $b(s)$.

\subsubsection{Gaussian Process UCB Optimization}
\label{gp_ucb_section}

The UCB algorithm described in Section \ref{rl_value} has been further developed to better handle the exploration-exploitation trade off in a sample efficient manner. 

\textbf{Context-free setting} One of the well-studied extensions to UCB in the context-free setting is the Gaussian Process UCB (GP-UCB) algorithm \cite{srinivas2009gaussian}. The algorithm combines the use of Gaussian Processes (GP) and UCB as an acquisition function. In simple terms a GP is a distribution over functions, which in multi-armed bandits is used to approximate the distribution over reward functions. A GP over the reward function is characterized by the mean function $\mu_i(\mathbf{a})$ and covariance function (or kernel function) $k_i(\mathbf{a},\mathbf{a}^{\prime})$. The assumption with using a GP is that our reward function is sampled from a Gaussian distribution which can be expressed as

\begin{equation}
    R(\mathbf{a}) \sim GP(\mu_i(\mathbf{a}),k_i(\mathbf{a},\mathbf{a}^{\prime})),
    \label{gp}
\end{equation}

where $GP(\mu(\mathbf{a}), K(\mathbf{a}, \mathbf{a}^{\prime}))$ represents a GP with prior mean $\mu$ and covariance function $K$, and $\mathbf{a} = \pi(s)$. 

The mean and the covariance of the posterior of the GP distribution after observations $D_{1:i} = \{\pi(s)_{1:i},r_{1:i}\}$ is defined as follows

\begin{align}
\mu_i(\mathbf{a}) &= k_{i-1}(\mathbf{a})^T(\mathbf{K}_{i-1} + \sigma^2\mathbf{I})^{-1}r_i,\\
ki(\mathbf{a},\mathbf{a}^{\prime}) &= k(\mathbf{a},\mathbf{a}^{\prime}) - k_{i-1}(\mathbf{a})^T(\mathbf{K}_{i-1} + \sigma^2\mathbf{I})^{-1}k_{i-1}(\mathbf{a}^{\prime}),\\
    \sigma^2_i(\mathbf{a}) &= k_i(\mathbf{a},\mathbf{a}^{\prime}),
\end{align}

where $\mu_i(\mathbf{a})$ is the posterior mean function, $k_i(\mathbf{a}, \mathbf{a}^{\prime})$ is the posterior covariance function, $k_{i-1}(\mathbf{a}) = [k(\mathbf{a}_1,\mathbf{a}),...,k(\mathbf{a}_T,\mathbf{a})]^T$ and $\mathbf{K}_T = [k(\mathbf{a},\mathbf{a}^{\prime})]_{\mathbf{a},\mathbf{a}^{\prime} \in A}$ (positive semi-definite covariance matrix), $r_i = R(\pi_i) + \epsilon_i$, and $\epsilon_i$ is modeled as a normally distributed noise. Given a GP approximation of the reward function from observations, the action-selection algorithm of GP-UCB is defined as follows

\begin{equation}
    A_i = {\arg\max}_{\mathbf{a}} \mu_i(\mathbf{a}) + \sqrt{\beta_i}\sigma_{i-1}(\mathbf{a}),
    \label{gp_ucb_eq}
\end{equation}

where $\beta_i = 2\log(|D|i^2\pi^2/6\delta)$ is a time-varying parameter that controls the degree of exploration. $|D|$ is the dimension of the action space, $\delta \in (0,1)$, and $\mu_{i-1}$ and $\sigma_{i-1}$ are the GP posterior mean and standard deviation influenced by samples collected before episode $i$.

The algorithm selects actions by choosing actions with high likelihood of having an uncertain reward function value (that is, where we have high variance $\sigma_{i-1}(\mathbf{a})$) and where we are reasonably certain that the action would result in a high reward function value (that is, where we have high mean $\mu_i(\mathbf{a}))$. A good choice of covariance function and proper choice of $\beta_i$ is needed for the algorithm to perform well (see Section \ref{gp_hypers} in the appendix for an analysis).

\textbf{Contextual setting} The GP-UCB algorithm has been extended into the contextual setting by introducing a collection of contexts $z$ to help the agent choose actions effectively in more situations. The extension is termed Contextual Gaussian Process UCB (CGP-UCB). The action-selection strategy is defined as below

\begin{equation}
    A_i = {\argmax}_{\mathbf{a}} \mu_i(\mathbf{a},z_t) + \sqrt{\beta_i}\sigma_{i-1}(a,z_t),
\end{equation}
where $\mu_{i-1}$ and $\sigma_{i-1}$ are as defined previously above, with the subtle difference that they are now defined on a set $X = A \times Z$ where $A$ and $Z$ are the finite action and context spaces respectively. Similar to the standard UCB, the state of the Malaria environment is used as context.

As more data (actions $\mathbf{a}_t$, context $z_t$ and noisy immediate rewards $R_t$) is collected, the algorithm with a given context $z_t$ uses the predictive power of the posterior GP to generate the mean and standard deviation to be used to select an action $\mathbf{a}_t$ and this gets better in the long run with a correct choice of kernel function (see Section \ref{gp_hypers} in the appendix).

\subsection{Black Box Optimization}
\label{optimization_baseline}

This section describes a class of machine-learning-based optimization methods that solve Malaria control as the general optimization problem \eqref{malaria_problem}, where the objective function $f$ has no known structure such as concavity and linearity, and potentially has no first-order or second-order derivatives. And since the policy (input vector $\mathbf{x}_i$) has no sequential structure, they also lack a mechanism to integrate state information. We use these methods as baselines.

In this work we implemented a random baseline (Section \ref{random_baseline}) and 2 popular optimization techniques, Genetic Algorithm (Section \ref{ga_section}) and Bayesian optimization (Section \ref{bayes_section}). We implemented these baselines in the context-free and contextual formulations\footnote{They are not applicable to MDPs because they do not use state information}. In the context-free setting, the policy $\mathbf{x}_i$ consists of a single 2-dimensional action $\mathbf{a}_i$, and for discrete actions it consists of a single one-dimensional discrete action. In the contextual setting, the policy $\mathbf{x}_i$ is a 10-dimensional vector consisting of 5 2-dimensional actions $\mathbf{a}_t$ as discussed in Section \ref{malaria_form}. Similarly, for discrete actions this is a 5-dimensional vector consisting of 5 one-dimensional discrete actions.

\subsubsection{Random Baseline}
\label{random_baseline}

The random algorithm is the simplest algorithm. Policies $\mathbf{x}_i$ are chosen randomly at every episode without considering any information from the environment, and the policy with the highest reward $r_i$ is returned: i.e., $\mathbf{x}_{\star} = {\argmax}_{\mathbf{x}} f(\mathbf{x})$. In the context-free setting, the policy $\mathbf{x}_i$ consists of a random action $\mathbf{a}_i$ that is repeatedly applied every year in each episode. In the contextual formulation the policy generates random actions $\mathbf{a}_t$ for each year. This baseline makes no assumptions about the structure of the environment beyond the dimensionality of the state and action spaces in each formulation. 

\subsubsection{Genetic Algorithm}
\label{ga_section}

A genetic algorithm (GA) is any population-based model inspired by Darwin's theory of evolution. 
GAs use selection and recombination operators (e.g., mutation and crossover) to generate new policies $\mathbf{x}_i$ in a search space (policy space $\Pi$). 
A population of candidate policies to \eqref{malaria_problem} is evolved towards better policies. 
Candidate policies are encoded into some problem-specific representation with certain properties (known as chromosomes or genotypes) which can be mutated and altered. 
Traditionally, a binary representation is used, but other encodings for discrete and continuous problems have been developed \cite{haupt2004practical}. 

A GA algorithm is generally composed of two components: a representation of the solution domain and an evaluation function. 
In Malaria control, we use either the discrete or continuous representations of the policy space. 
The evaluation function is the reward $r_i$. 
The evolution to the best policies starts from a population of randomly generated policies (called a generation), and in each iteration the fitness of every policy is evaluated. 
The best policies with respect to the evaluation $r_i$ are selected with some probability, and are recombined and mutated to form a new generation of policies. 
In addition, in the Elitist variants, the best policies from the current generation are carried over to the next, unaltered, to guarantee that the policy quality does not degrade catastrophically from one generation to the next. 

The new generation of policies is then used in the next iteration of the algorithm. 
The algorithm terminates when either the maximum number of generations has been produced, or a satisfactory fitness level has been reached for the population. 
Here, we terminate the algorithm when the maximum number of function evaluations reaches a threshold\footnote{This determines the maximum number of candidate policies created and evaluated}.

\subsubsection{Bayesian Optimization}
\label{bayes_section}

Bayesian optimization (Bayes Opt) makes further assumptions on the optimization problem \eqref{malaria_problem}. 
It assumes that the policy space $\Pi$ is a simple set, in which it is easy to evaluate membership, and that the reward function $r_i$ is continuous and noisy.
It is particularly useful when the reward function is expensive to evaluate, because it minimizes the number of function evaluations to find the optimal value. 

It consists of two main components: a regressor, $\hat{f}(\mathbf{x})$, (also known as a surrogate function) for modeling the objective function $f(\mathbf{x})$ (usually a statistical model such as a Gaussian process \eqref{gp}), and an acquisition function $u(\mathbf{x})$ for sampling the next policy $\mathbf{x}_i$. The Bayes Opt algorithm starts with an initial set of policies, covering the policy space, usually sampled uniformly at random, and uses them to iteratively allocate the remainder of a budget of $N$ function evaluations, or episodes. 
The statistical model provides a Bayesian posterior distribution that describes reward values$\mathbf{x}_i$ for a candidate policy $\mathbf{x}_i$. 
The posterior distribution is updated at each episode $i$ after observing $r_i$ at a new policy $\mathbf{x}_i$.

The acquisition function estimates the reward that would be generated if a candidate policy $\mathbf{x}_i$ was executed, based on the current posterior distribution over $f(\mathbf{x})$. 
There are several different acquisition functions in the literature, which is often difficult to choose in practice. This includes expected improvement (EI), probability of improvement (PI), knowledge gradient (KG), UCB, GP-hedge, etc. 
We use the GP-hedge function, which probabilistically chooses amongst three acquisition functions, UCB, EI and PI \cite{hoffman2011portfolio}. 
Each acquisition function is optimized independently to propose a candidate policy $\mathbf{x}_i^j = \argmax_{\mathbf{x}}u_j(\mathbf{x})$. 
The next candidate policy for evaluation $\mathbf{x}_i$ is chosen out of the three with probability

\begin{equation}
    p_i(j) = \frac{e^{\eta g^j_{t-1}}}{\sum_{b=1}^{k} e^{\eta g^b_{t-1}}},
    \label{gp_hedge}
\end{equation}
where $k=3$ is the number of individual acquisition functions, $g^j_{i-1}$ is a weight for $u_j(\mathbf{x})$ and is initialized to zero and updated every episode using $g^j_i = g^j_{i-1} + \mu_i(\mathbf{x}_t^j)$ and $\eta$ is a parameter. 
$\mu_i$ is the mean function of the newly updated GP regressor $\hat{f}$. 
The reader is referred to \cite{hoffman2011portfolio,frazier2018tutorial} for a detailed treatment.

\section{Experiments}
\label{experiments}

This section presents the analysis of the algorithms described in Section \ref{algorithms}, applied to settings discussed in Section \ref{malaria_form} using experiments ran on the OpenAI gym Malaria environment \cite{pmlr-v123-remy20a}. 
We analyze results of the algorithms in each setting separately in Section \ref{context-free-results} for Context-free, Section \ref{contexual-results} for Contextual and Section \ref{mdp-results} for MDP. 
We also analyze results across the different settings in Section  \ref{all-results}. 
Finally, we conclude this section with a discussion of our analysis in Section \ref{discussion-results}.

We follow the same experimental setup as described in the Live Malaria Challenge 2019 \footnote{https://compete.hexagon-ml.com/practice/rl\_competition/38/} \cite{pmlr-v123-remy20a}. 
Each algorithm is limited to a maximum of 2000 environmental steps to learn a policy. 
This simulates the limited number of evaluations on real epidemiological models as they can be expensive to compute in practice\footnote{Running one instance of an OpenMalaria simulation, for a representative human population size (100,000), can return results in the time frame of hours \cite{bent2018novel}}, where sample efficient algorithms are required. 
Since each episode of the environment consists of 5 steps (see Figure \ref{MalariaGym}), each algorithm has a maximum of 400 episodes. 

We use the first 399 episodes for training and the last one for evaluation. 
Each experiment is repeated 20 times for each algorithm and we analyze the learning behavior of algorithms in terms of sample efficiency and the asymptotic performance of the learned policies. 
We defer the discussion of hyperparemeter tuning to Appendix \ref{hyperparameters}, and only report results for the best hyperparameters. 

\subsection{Context-free setting}
\label{context-free-results}

\begin{figure}[t!]
    \centering
    \begin{subfigure}[t]{0.5\textwidth}
        \centering
        \includegraphics[height=1.7in]{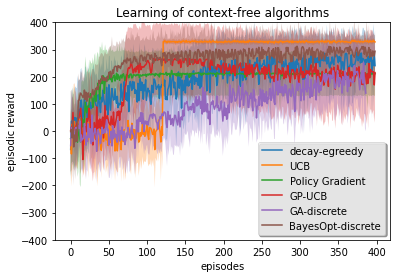}
        \caption{Training}
    \end{subfigure}%
    ~ 
    \begin{subfigure}[t]{0.5\textwidth}
        \centering
        \includegraphics[height=1.7in]{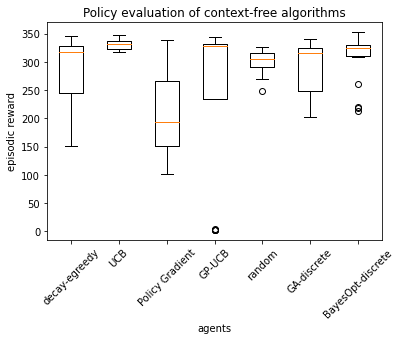}
        \caption{Policy evaluation}
    \end{subfigure}
    \caption{Comparison of algorithms in the Context-free setting.}
    \label{context_free_results}
\end{figure}

Figure \ref{context_free_results} shows a comparison of the algorithms in the context-free setting.
The algorithms applicable to this setting include $\epsilon$-greedy, UCB, policy gradient, GP-UCB and the baselines, random, Genetic Algorithm (GA) and Bayesian Optimization (Bayes Opt).

\textbf{Training} Amongst the RL algorithms, the policy gradient algorithm learns faster in earlier episodes ($i < 70$), followed by $\epsilon$-greedy and then GP-UCB, but all of them quickly converged below an average episodic reward of 200 while $\epsilon$-greedy converged just below 300. 

UCB performed poorly in early episodes but suddenly improved and converged immediately above 300. 
A closer look reveals that the sudden improvement occurs exactly at episode 121, coinciding with the number of discrete actions available (see \eqref{discrete_actions}). 
This shows that UCB (see \eqref{ucb_original}) initially explores all discrete policies at least once to discover a good set of policies with high rewards $r_i$ and starts exploiting these as their estimated values $Q_i(\mathbf{a})$ outweigh the exploration term $c\sqrt{\frac{\ln{i}}{N_i(\mathbf{a})}}$.
This suggests that in the context-free setting, there is a set of actions that, if applied every year, provide the highest reward and UCB is able to find these actions consistently.   

This experiment also supports the claim that a Gaussian Process extension of UCB is more sample efficient than the standard UCB \cite{srinivas2009gaussian}, as it learns to achieve higher rewards earlier. 
Note that if evaluations were limited, to say 100 episodes, then GP-UCB would learn a better policy than the others and UCB would fail to learn a good policy.   

\textbf{Policy evaluation} Only policies learned by UCB outperform the random baseline on average. 
The policies learned by GP-UCB have a larger variance compared to UCB and its learning performance degrades over time. 
While the asymptotic performance of Bayes Opt is inferior to UCB, its learning is more sample efficient than that of all other implemented methods.

\subsection{Contextual setting}
\label{contexual-results}

\begin{figure}[t!]
    \centering
    \begin{subfigure}[t]{0.5\textwidth}
        \centering
        \includegraphics[height=1.7in]{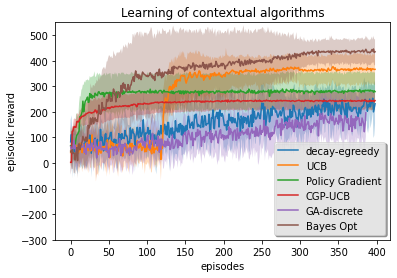}
        \caption{Training}
    \end{subfigure}%
    ~ 
    \begin{subfigure}[t]{0.5\textwidth}
        \centering
        \includegraphics[height=1.7in]{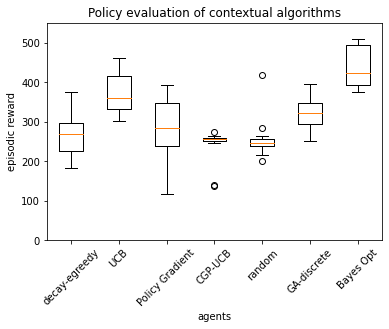}
        \caption{Policy evaluation}
    \end{subfigure}
    \caption{Comparison of algorithms in the Contextual setting.}
    \label{contextual_results}
\end{figure}

Figure \ref{contextual_results} shows a comparison of the same algorithms applied in the contextual setting.

\textbf{Training} It is slightly harder for policy gradients and decay $\epsilon$-greedy to learn initially because the search space is larger compared to the context-free setting.
In fact, decay $\epsilon$-greedy performs worse than in the context-free setting, converging around an average reward of 200. 
Policy gradient eventually performs better than decay $\epsilon$-greedy and CGP-UCB by a large margin, converging around an average reward of 300. As in the context-free setting, UCB starts slow but suddenly improves at episode 121 and converges at a higher average reward of around 350. 
It converges slower than in the context-free setting and also has a higher variance in its performance. 
This is most likely due to the fact that it now has to optimize for a sequence of actions rather than a single action, which increases the size of the search space.

The sample efficiency of the Gaussian Process extension of UCB is even better in the contextual setting as it achieves higher rewards in earlier episodes. 
However, CGP-UCB quickly converges to an inferior policy compared to UCB and policy gradients. 
The results suggest that in general the contextual setting allows the discovery of better policies, with the exception of the decay $\epsilon$-greedy algorithm.

\textbf{Policy evaluation} Policies learned by UCB, GA (discrete) and Bayes Opt significantly outperform the random baseline. 
The random baseline performs poorly compared to the context-free setting.
This is because randomly sampling a higher dimensional action space in search for an optimum is much harder. 
Thus, this increases the need for more intelligent sampling of the search space.
The contextual $\epsilon$-greedy, policy gradients and GA fail to outperform the context-free random baseline. 

The learning of UCB is less consistent than in the context-free setting but the policies learned perform better on average, sometimes achieving rewards above 400.
Contextual UCB performs better than all baselines except for Bayesian optimization.
Bayesian optimization achieves rewards above 400 on average, sometimes reaching rewards of 500. 
This also supports the notion that learning to predict actions separately for each year (context) may be better than a single action, which is generally done in the Malaria epidemiology literature \cite{bent2018novel,korenromp2016malaria}.

\subsection{Markov Decision Process setting}
\label{mdp-results}

\begin{figure}[t!]
    \centering
    \begin{subfigure}[t]{0.5\textwidth}
        \centering
        \includegraphics[height=1.7in]{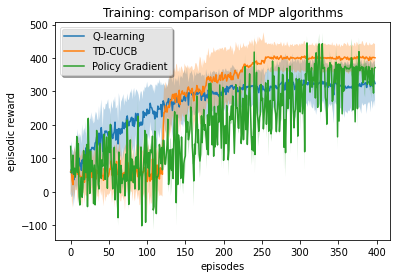}
        \caption{Training}
    \end{subfigure}%
    ~ 
    \begin{subfigure}[t]{0.5\textwidth}
        \centering
        \includegraphics[height=1.7in]{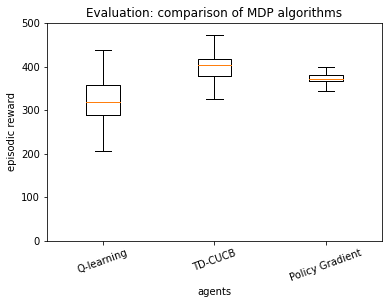}
        \caption{Policy evaluation}
    \end{subfigure}
    \caption{Comparison of algorithms in the MDP setting.}
    \label{mdp_results}
\end{figure}

Figure \ref{mdp_results} shows a comparison of algorithms applied in the MDP setting, which includes Q-learning with decay $\epsilon$-greedy and UCB (TD-CUCB), and Policy gradients. 

\textbf{Training} Policy gradient learns slower than the rest but converges to average rewards higher than Q-learning and is more consistent. 
The learning behavior of TD-CUCB is the same as in the context-free and contextual settings, as it suddenly improves at episode 121. 
TD-CUCB has a better asymptotic performance than all others. 

\textbf{Policy evaluation} The learned policies of TD-CUCB perform better than the rest but those of the Policy gradient are more consistent. 
However, while the performance of TD-CUCB varies more, it sometimes achieves rewards as high as 500. Q-learning fails to outperform the context-free random baseline.

\subsection{All settings}
\label{all-results}

\begin{figure}[t!]
    \centering
    \begin{subfigure}[t]{0.5\textwidth}
        \centering
        \includegraphics[height=1.7in]{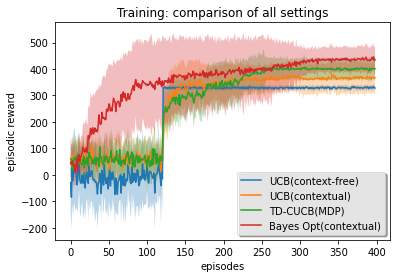}
        \caption{Training}
    \end{subfigure}%
    ~ 
    \begin{subfigure}[t]{0.5\textwidth}
        \centering
        \includegraphics[height=1.7in]{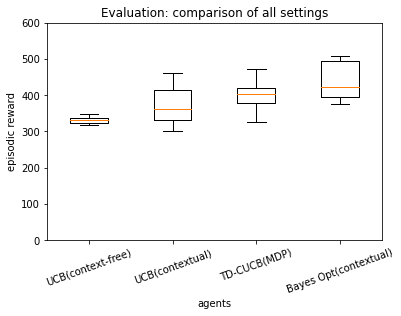}
        \caption{Policy evaluation}
    \end{subfigure}
    \caption{Comparison of best algorithms in all settings.}
    \label{all_results}
\end{figure}

Figure \ref{all_results} shows a comparison of the best performing RL algorithms across the different settings, which are all UCB-based, and the best performing baseline, Bayes Opt.
As already described in Section \ref{contexual-results}, the context-free formulation restricts learning to a single action $\mathbf{a}_i$, repeatedly applied every year. 
TD-CUCB in the MDP formulation performs slightly better than UCB in the contextual formulation, achieving rewards above 500 at times.
There is an increasing trend in performance from context-free to contextual and MDP settings, suggesting that actions are correlated across states, and that context-free and contextual formulations do not capture all important aspects of the Malaria environment.
Overall, the RL algorithms were unable to perform better than Bayesian optimization.

\subsection{Discussion}
\label{discussion-results}

We perform ablation studies on the effect of hyperparemters on the performance of each algorithm in the Appendix (Section \ref{hyperparameters}) and also analyze each algorithm separately across the  different formulations in Section \ref{all_algos_settings}. 
The experiments show that all RL algorithms in all formulations are affected by the choice of hyperparameters, with Policy gradients in the MDP setting (REINFORCE) -- the only algorithm to use a neural network policy -- exhibiting the highest variance in performance. 
They also show that, amongst the RL agents, standard UCB consistently performs the best across formulations.
Furthermore, the fact that it only has one hyperparameter to tune ($c$ in the exploration term of \eqref{ucb_original}) may be appealing to practitioners\footnote{All RL algorithms also have the learning rate as a hyperparameter but we used a fixed value of 0.9.}. 
More importantly, this value does not seem to have a significant impact on the asymptotic performance of UCB (see Figure \ref{ucb_hyper} in the Appendix).

Bayes Opt performs well, probably due to the low dimensional search space (10 variables) but would probably struggle with a problem with many intervention choices over a longer timescale. 
However, this suggests that a Bayesian approach (which actively samples the search space and is thus sample efficient) and a sequential algorithm (with an MDP formulation, that reduces the search space) may be ideal in most cases.

\section{Conclusion}
\label{conclusion}

This work studies the formulation of Malaria control, a problem of finding optimal interventions to reduce the mortality and morbidity rate of malaria, and provides a comprehensive analysis of reinforcement learning and black box optimization algorithms on three formulations of the problem. 
While the majority of work on Malaria control uses one of the formulations and focuses on developing new algorithms (or implementing some of the advanced RL algorithms), this work conducts experiments to show which formulation is best suited for the problem, and shows that standard RL algorithms such as UCB perform competitively.

Q-learning with UCB exploration in the MDP formulation, which we termed TD-CUCB, outperformed all other RL algorithms across all formulations.
However, none of the RL algorithms was able to outperform Bayesian optimization, which proved to be both sample efficient and learned superior Malaria policies in the contextual formulation.
Nevertheless, our experiments show that simple algorithms coupled with a formulation that captures correlations between states and actions are sufficient to learn good Malaria policies. 
While Bayesian optimization, like many black box optimization algorithms, does not capture the sequential nature of the problem, it accounts for interactions between actions and, coupled with the low-dimensionality of the action space, is able to search for good policies efficiently.


\appendix
\section*{Appendix A. Ablation Studies}

\subsection{Hyperparameters}
\label{hyperparameters}

We report hyperparameter analysis for all algorithms in each formulation. The random baseline has no hyperparameter. For Genetic Algorithm we used hyperparameters shown in Table \ref{ga_hyperparams}. We set the population size and the maximum number of iterations such that, given default values for the rest\footnote{We used the GA library: https://pypi.org/project/geneticalgorithm/\#1111-id}, the total number of function evaluations does not exceed 399 episodes.  

For Bayesian optimization we use the Scikit optimize library\footnote{https://github.com/scikit-optimize/scikit-optimize} and used default parameters as shown in Table \ref{bo_hyperparams} and only change the number of function evaluations ($n\_calls$) to 399. For GA and Bayesian optimization baselines, we experiment with both continuous and discrete action spaces using the same hyperparameters.

\begin{table}[htbp]
\centering
\catcode`,=\active
\def,{\char`,\allowbreak}
\renewcommand\arraystretch{1.2}
\begin{tabular}{p{3.5cm}<{\raggedright} p{3.5cm}}
  \toprule
    Hyperparameters           & \textbf{Values}                 \\ 
  \midrule
    n\textunderscore calls   & 399                     \\
    n\textunderscore initial \textunderscore points &  10\\
    acq\textunderscore func     & gp\textunderscore hedge \\
    kappa &  1.96\\
    n\textunderscore points    & 10000                         \\
  \bottomrule
\end{tabular} 
\caption{Hyperparameters for Bayesian optimization baseline.}
\label{bo_hyperparams}
\end{table}

In the following sections we analyze hyperparameters for the learning algorithms in all formulations. All algorithms except for GP-UCB and CGP-UCB use a fixed learning rate of $\alpha = 0.9$.

\subsubsection{$\epsilon$-greedy}

\begin{figure}[t!]
    \centering
    \begin{subfigure}[t]{0.5\textwidth}
        \centering
        \includegraphics[height=1.7in]{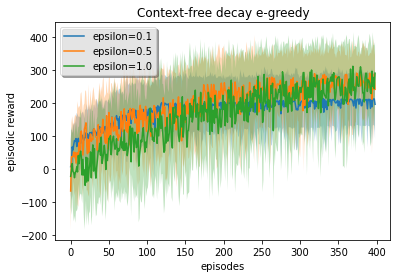}
        \caption{Context-free}
    \end{subfigure}%
    ~ 
    \begin{subfigure}[t]{0.5\textwidth}
        \centering
        \includegraphics[height=1.7in]{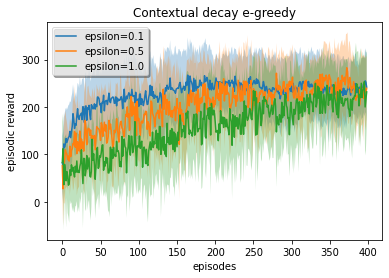}
        \caption{Contextual}
    \end{subfigure}
    ~ 
    \begin{subfigure}[t]{0.5\textwidth}
        \centering
        \includegraphics[height=1.7in]{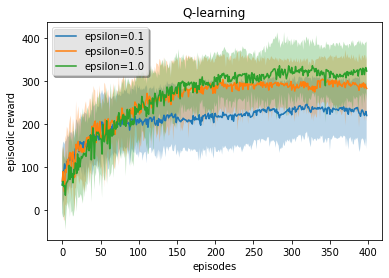}
        \caption{MDP}
    \end{subfigure}
    \caption{Comparison of hyperparameters for decay $\epsilon$-greedy in all formulations.}
    \label{e_greedy_hyper}
\end{figure}

Here we analyze hyperparameters for algorithms that use decay $\epsilon$-greedy in all formulations, with $\epsilon$ as the hyperparameter. We decay $\epsilon$ from some initial value (we try $0.1, 0.5, 1.0$) to the final value of $0.01$ over $200$ episodes. We chose a decay period of $200$ so that it uses half the time to explore the actions but different values do not improve results significantly. Figure \ref{e_greedy_hyper} shows a comparison of the hyperparameters across the formulations. $\epsilon = 1.0$ performs better consistently across the formulations but it is not statistically significant. Notably, in the contextual formulation, $\epsilon = 0.1$ is more sample efficient but still has the same asymptotic performance as others. So we use $\epsilon = 1.0$ in the main experiments.

\begin{table}
\centering
\catcode`,=\active
\def,{\char`,\allowbreak}
\renewcommand\arraystretch{1.2}
\begin{tabular}{p{4.5cm}<{\raggedright} p{2.5cm}}
  \toprule
    Hyperparameters           & \textbf{Values}                 \\ 
  \midrule
    Maximum number of iteration  &   5                    \\
    Population size      &  87\\
    Mutation probability     & 0.1 \\
    Elite ratio &  0.01\\
    Crossover probability      & 0.5                         \\
    Parents portion                 & 0.3                        \\ 
    Crossover type    & Uniform \\
    Maximum iteration without improvement & None\\
  \bottomrule
\end{tabular} 
\caption{Hyperparameters for Genetic Algorithm baseline.}
\label{ga_hyperparams}
\end{table}

\subsubsection{UCB}

\begin{figure}[t!]
    \centering
    \begin{subfigure}[t]{0.5\textwidth}
        \centering
        \includegraphics[height=1.7in]{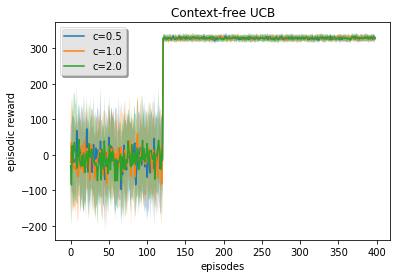}
        \caption{Context-free}
    \end{subfigure}%
    ~ 
    \begin{subfigure}[t]{0.5\textwidth}
        \centering
        \includegraphics[height=1.7in]{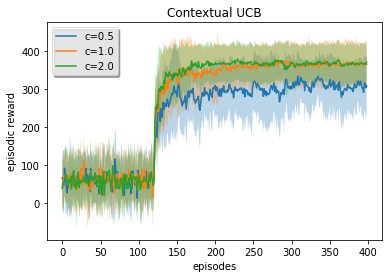}
        \caption{Contextual}
    \end{subfigure}
    ~ 
    \begin{subfigure}[t]{0.5\textwidth}
        \centering
        \includegraphics[height=1.7in]{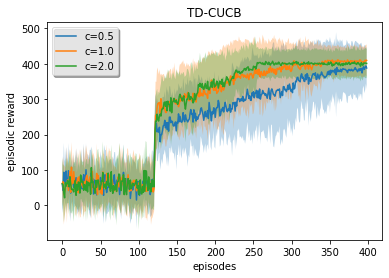}
        \caption{MDP}
    \end{subfigure}
    \caption{Comparison of hyperparameters for UCB in all formulations.}
    \label{ucb_hyper}
\end{figure}

Here we analyze hyperparameters for algorithms that use the standard UCB in all formulations. The standard UCB only has one hyperparameter, $c$, which aims to balance between exploration and exploitation (see Eq. \ref{ucb_original} and \ref{td_cucb}). Figure \ref{ucb_hyper} shows a comparison across the formulations. $c = 2$ consistently performs better or the same as other values across the formulations. So we use $c = 2$ in the main experiments. These results also show that choosing a different value would not hurt the asymptotic performance significantly in cases where one is not able to optimize for the best value, but larger values are preferable for sample efficiency.

\subsubsection{GP-UCB and CGP-UCB}
\label{gp_hypers}

As described in Section \ref{gp_ucb_section}, GP-UCB is governed by the choice of $\beta_{t}$ (Eq. \ref{gp_ucb_eq}), the kernel function $K$ for the GP (Eq. \ref{gp}) and the Gaussian noise $\epsilon_t$ on the observations. Following the work of \cite{bent2018novel}, we choose the Matern-$5/2$ kernel, which is governed by the length scale $l$ and the smoothness parameter $\nu = \frac{5}{2}$. We also experimented with a Radial Basis Function (RBF) kernel but it resulted in similar performance to the Matern kernel, so we do not report it here for brevity. 

\begin{figure}[t!]
    \centering
    \begin{subfigure}[t]{0.5\textwidth}
        \centering
        \includegraphics[height=1.7in]{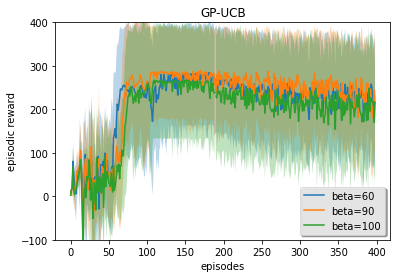}
        \caption{GP-UCB with Matern kernel}
    \end{subfigure}%
    ~ 
    \begin{subfigure}[t]{0.5\textwidth}
        \centering
        \includegraphics[height=1.7in]{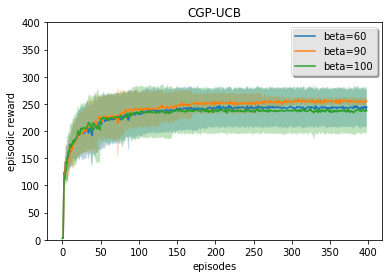}
        \caption{CGP-UCB composed of RBF (contexts) and Matern (actions) kernels.}
    \end{subfigure}
    \caption{Comparison of hyperparameters for GP-UCB and CGP-UCB.}
    \label{gp_hyper_freq1}
\end{figure}

To speed up computation, we use the Sparse GP implementation\footnote{https://github.com/SheffieldML/GPy} with 100 inducing points. For a time-varying $\beta_t$ we experiment with $\delta = \{0.3, 0.6, 0.8\}$ and for each we optimize for $l$. We also experiment with fixed $\beta_t = \{60, 90, 100\}$. We found the performance when setting $\beta_t$ to fixed values to be more consistent and only report on this for brevity. Figure \ref{gp_hyper_freq1} (top) shows the performance of GP-UCB for different values of $\beta_t$. We set $l = 1.0$ as starting value and optimize it together with the GP model. The value of $\beta_t$ does not affect the performance of GP-UCB significantly. We use $\beta_t = 90$ in the main experiments.

CGP-UCB is governed by similar hyperparameters. We choose a kernel function composed of a product between a radial basis (RBF) kernel for the contexts and Matern-$5/2$ (with $l_{matern}$ as hyperparameter) for the actions \cite{NIPS2011_f3f1b7fc}. This introduces additional hyperparameters due to the RBF kernel, which are the variance $\sigma_{rbf}$ and the length scale $l_{rbf}$. We also experimented with an RBF kernel for both contexts and actions and it performed poorly, so we do not report this. We used the same $\beta_t$ values as with GP-UCB. A time-varying $\beta_t$ performed similar to using fixed values, so we do not report it. We also use a Sparse GP with 100 inducing points. 

Due to the complexity of training a GP model, we update the GP model every 10 iterations (2 episodes) to speed up training. Figure \ref{gp_hyper_freq1} (bottom) shows the performance of CGP-UCB for different values of $\beta_t$. We fixed $l_{matern}, \sigma_{rbf}, l_{rbf}$ to their default value of $1.0$. Surprisingly, optimizing these values resulted in the reward values collapsing to 0.

\subsubsection{Policy Gradient}

\begin{figure}[t!]
    \centering
    \begin{subfigure}[t]{0.5\textwidth}
        \centering
        \includegraphics[height=1.7in]{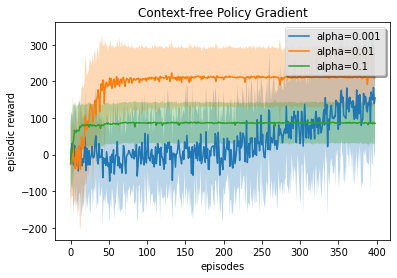}
        \caption{Context-free}
    \end{subfigure}%
    ~ 
    \begin{subfigure}[t]{0.5\textwidth}
        \centering
        \includegraphics[height=1.7in]{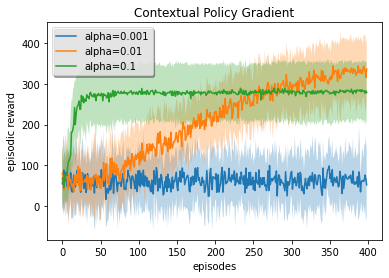}
        \caption{Contextual}
    \end{subfigure}
    ~ 
    \begin{subfigure}[t]{0.5\textwidth}
        \centering
        \includegraphics[height=1.7in]{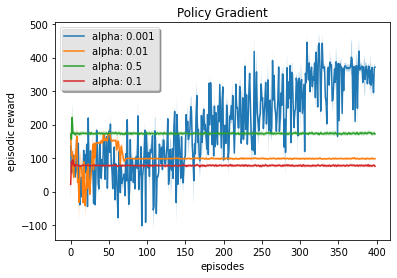}
        \caption{MDP}
    \end{subfigure}
    \caption{Comparison of hyperparameters for policy gradients in all formulations.}
    \label{pg_hyper}
\end{figure}

Here we analyze hyperparameters for policy gradients in all formulations. The only hyperparameter to tune for policy gradients in the Bandit settings is the learning rate. The REINFORCE algorithm uses a neural network to parameterize the policy so it has additional hyperparameters. We used a neural network with one hidden layer and used grid search to optimize the number of neurons (we try $6, 10, 20, 50, 200, 500$) and the learning rate (we try $0.001,0.01,0.1,0.5$). We only report results for 10 hidden layers for brevity as this performed better. Figure \ref{pg_hyper} shows a comparison of the hyperparameters across the formulations. 

The results show that small learning rates are preferable for REINFORCE and too small or too large for Bandits is detrimental. So in the main experiment we use a learning rate of 0.01 in the Bandit settings, 0.001 in the MDP setting and a neural network with 10 hidden neurons. 

\subsection{Analysis of Algorithms across settings}
\label{all_algos_settings}

\begin{figure}[h]
    \centering
    \begin{subfigure}[t]{0.4\textwidth}
        \centering
        \includegraphics[height=1.7in]{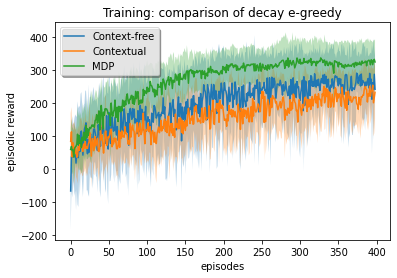}
        \caption{decay $\epsilon$-greedy}
    \end{subfigure}%
    ~ 
    \begin{subfigure}[t]{0.4\textwidth}
        \centering
        \includegraphics[height=1.7in]{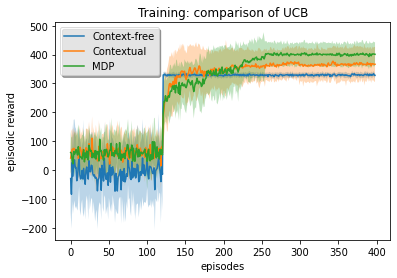}
        \caption{standard UCB}
    \end{subfigure}
    ~ 
    \begin{subfigure}[t]{0.4\textwidth}
        \centering
        \includegraphics[height=1.7in]{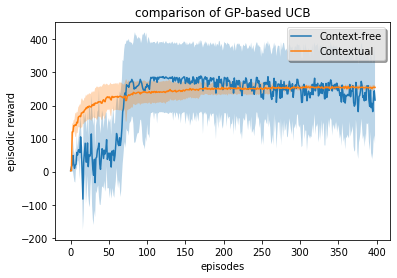}
        \caption{GP-based UCB}
    \end{subfigure}
    ~ 
    \begin{subfigure}[t]{0.4\textwidth}
        \centering
        \includegraphics[height=1.7in]{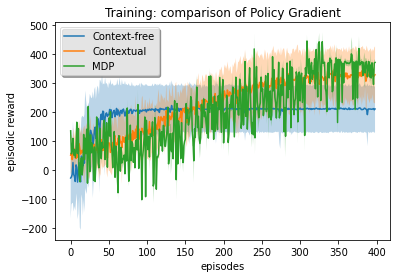}
        \caption{Policy gradient}
    \end{subfigure}
    \caption{Comparison of learning algorithms in all formulations.}
    \label{across_settings}
\end{figure}

In this section we analyze the performance of each algorithm separately across formulations. The aim is to further highlight the effect that each formulation has on the performance of the algorithms. We use the best parameters for each algorithm as reported in Section \ref{hyperparameters}. Figure \ref{across_settings} shows a comparison of each learning algorithm across the formulations. UCB and policy gradients perform better in contextual and MDP settings. The performance of policy gradients between contextual and MDP is not significantly different from each other but UCB performs significantly better in the MDP formulation. Perhaps surprisingly, the decay $\epsilon$-greedy performs worse in the contextual formulation than in the context-free setting. However, consistent with the other algorithms, it performs better in the MDP formulation.

\begin{figure}[t]
    \centering
    \begin{subfigure}[t]{0.5\textwidth}
        \centering
        \includegraphics[height=1.7in]{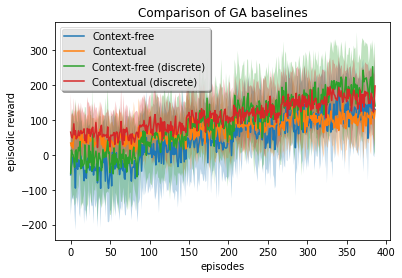}
        \caption{Genetic Algorithm}
    \end{subfigure}%
    ~ 
    \begin{subfigure}[t]{0.5\textwidth}
        \centering
        \includegraphics[height=1.7in]{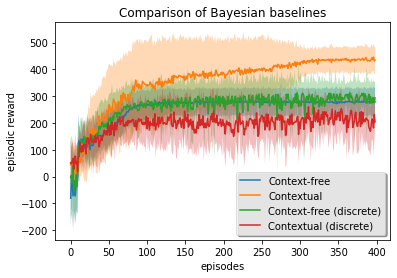}
        \caption{Bayesian optimization}
    \end{subfigure}
    \caption{Comparison of baselines in all formulations.}
    \label{across_baselines}
\end{figure}

Figure \ref{across_baselines} shows a comparison of the baselines across the formulations. The random baseline struggles to achieve rewards that are significantly above zero on average, with the contextual formulation only helping it achieve a lower variance. For GA and Bayes Opt we conducted experiments for both continuous and discrete (using the same discretization in Equation \ref{discrete_actions}) action spaces. Discretization seems to improve the performance of GA (perhaps due to reduced action space) while Bayesian optimization performs better in the continuous action space. Overall, the contextual formulation provides the best setting for the baselines to perform well, with Bayesian optimization outperforming all by a large margin.

\vskip 0.2in
\bibliography{main}
\bibliographystyle{theapa}

\end{document}